\documentclass[submission,copyright,creativecommons]{eptcs}
\providecommand{\event}{ACT 2026} 

\usepackage{iftex}
\usepackage{amsmath}
\usepackage{amssymb}
\usepackage{amsthm}
\usepackage{caption}
\usepackage{booktabs}
\usepackage{subcaption}

\usepackage[ruled,vlined]{algorithm2e}

\usepackage{tikz}
\usepackage{tikz-cd}
\usetikzlibrary{shapes, arrows, positioning, calc, fit}

\tikzset{
    >=stealth',
    rect/.style={rectangle, draw, minimum size=0.8cm}, 
}

\ifpdf
  \usepackage{underscore}         
  \usepackage[T1]{fontenc}        
\else
  \usepackage{breakurl}           
\fi

\newtheorem{definition}{Definition}[section]

\title{Categorical Internalisation of Environmental Groupoids for Generalisable POMDP Solving}
\def\titlerunning{Categorical Internalisation of Environmental Groupoids for Generalisable POMDP Solving}

\author{Ben Opperman
\institute{\scriptsize City St George’s, University of London}
\email{\scriptsize Ben.Opperman@citystgeorges.ac.uk}
\and
Eduardo Alonso
\institute{\scriptsize City St George’s, University of London}
\email{\scriptsize E.Alonso@citystgeorges.ac.uk}
\and
Esther Mondragón
\institute{\scriptsize City St George’s, University of London}
\email{\scriptsize E.Mondragon@citystgeorges.ac.uk}
}
\def\authorrunning{Opperman, Alonso \& Mondragón}

\begin{document}
\maketitle

\begin{abstract}
    This paper advocates category theory as a practical framework for structuring and improving reinforcement learning in high-dimensional, partially observable environments. We model symmetries between environmental states by partitioning the state space into equivalence classes induced by symmetry orbits, and organise each such class as a groupoid with a designated canonical representative. This allows the agent to share what it learns across many similar environmental states simultaneously, rather than treating every orientation or position as an entirely new problem. Learning is thus carried out on a symmetry-reduced state space with each orbit represented once, preserving structure while eliminating redundancy and improving sample efficiency.
    
    We implement this framework within standard reinforcement learning pipelines and evaluate two different approaches on partially observable benchmarks, demonstrating that orbit-based partitioning yields consistent performance improvements in environments exhibiting latent symmetry. Beyond these empirical results, our approach illustrates how categorical structure provides a principled bridge between abstract reinforcement learning formulations and their computational application, thereby establishing a pathway toward more structured and scalable learning systems.
\end{abstract}

\section{Introduction}
Reinforcement learning (RL) has seen widespread success in solving complex sequential decision-making problems, with myriad applications from board games to robotic control systems \cite{silver2017mastering, levine2016end}. In the standard framework, an agent interacts with an environment modelled as a Markov Decision Process (MDP), using reward information from its interactions to refine a policy that maximizes a cumulative reward \cite{sutton2018reinforcement, puterman1994markov}. However, RL approaches have faced some challenges resolving real-world, high-dimensional environmental complexity, particularly with sample inefficiency and poor generalisation \cite{henderson2018deep}.

Category theory holds potential to redefine environmental representations for RL agents, and this paper will utilise it as a foundation to accelerate learning by exploiting environmental symmetries. From a categorical perspective, an MDP can be understood as a coalgebra for the Giry monad, where transitions represent morphisms between state objects \cite{giry1982categorical, jacobs2016coalgebra}. Given a measurable state space $X$ and a set of actions $A$, the system dynamics are given by a measurable map $P : X \times A \to \mathcal{G}(X)$, where $\mathcal{G}(X)$ denotes the space of probability measures over $X$. In practical applications, agents lack complete information and instead of directly accessing $X$ they experience a partially observable problem with data given through high-dimensional, often-noisy observations. Thus, maintaining a robust internal representation which properly captures the valuable information is vital and for an agent to generalise across disparate states, it must learn to map raw observations into a latent space that disentangles the underlying factors of variation \cite{bengio2013representation, higgins2018definition}. Disentanglement allows the agent to treat independent properties of the world such as position, color, or velocity, as distinct coordinates in its internal model, creating stronger understandings about the environment while limiting weak conclusions drawn from few samples \cite{rahaman2020extrapolation}.

One key pathway to effectively achieve  disentanglement is by utilising environmental symmetries, the properties inherently invariant under various transformation groups, such as translations or rotations \cite{noether1918invariante}. By incorporating these symmetries into the agent's architecture, an agent can reduce its hypothesis space to augment sample efficiency and support shared learning across different but symmetrical states \cite{cohen2016group, vanderpol2020mdphom}. In this paper, we propose \textit{groupoids} as the natural mathematical language for internal representations in RL, and we show how to use them to capture the local structure of reversible actions and substructure transitions, as has been done in other fields before \cite{weinstein1996groupoids, vistoli2005groupoids}. This groupoid-based approach both generalises existing symmetry-based RL methods and also provides a compositional framework for building complex internal models from simpler geometric primitives \cite{opperman2026groupoid}. 

This paper will outline a novel RL framework which uses groupoids for internal representations to facilitate shared learning across states and accelerates learning in partially observable environments. Our contributions are as follows: \textbf{categorical reinforcement learning}: we show how category theory can be used to better represent environments for RL models; \textbf{groupoid solver algorithm}: we implement a practical algorithm using this categorical internal representation for the first time; and \textbf{empirical results}: we use standard benchmarks to demonstrate the performance of our model against traditional approaches.

In the following, we first review imperative background information to recontextualise reinforcement learning and symmetry-based methods using category theory. We will outline both the mathematical and then computational models for groupoid-based RL with different canonicalisation methods as well as the setup for the testing environments. Finally, we will analyse the impact of different approaches using key environmental benchmarks, and discuss the results and their implications.

\section{Background}
In this section, we reformulate standard RL in categorical terms to provide a rigorous basis for our model. We first establish environmental structure, and then outline categorical definitions of symmetry and disentanglement to create the bedrock for canonicalisation methods which reduce sample complexity.

\subsection{Reinforcement Learning}
RL agents learn through sequential interactions with an environment to maximize cumulative reward \cite{sutton2018reinforcement}. Most such problems are formalised as Markov Decision Processes (MDPs) \cite{puterman1994markov}, which can be lifted into a categorical framework to reveal their underlying algebraic structure.

\begin{definition}[Categorical Markov Decision Process]
A \textit{Markov Decision Process (MDP)} is a tuple $\mathcal{M} = (S, A, \alpha, \gamma)$ in the category of measurable spaces, where $S$ is the state space, $A$ is the action space,
\[
\alpha: S \times A \to \mathcal{G}(S \times \mathbb{R})
\]
is a Markov kernel assigning to each state-action pair a probability measure over successor states and rewards, and $\gamma \in [0,1)$ is the discount factor. Equivalently, by currying, $\alpha$ induces a morphism
\[
\tilde{\alpha} : S \to \mathcal{G}(S \times \mathbb{R})^A,
\]
making $S$ a coalgebra for the functor $F(X) = \mathrm{Meas}(A, \mathcal{G}(X \times \mathbb{R}))$.

The classical transition kernel $P$ and expected reward function $R$ are recovered via marginalisation and integration:
\[
P(s, a, B) = \alpha(s, a)(B \times \mathbb{R}), \quad
R(s, a) = \int_{S \times \mathbb{R}} \pi_2 \, d\alpha(s, a),
\]
where $\pi_2 : S \times \mathbb{R} \to \mathbb{R}$ is the projection onto the reward component.
\end{definition}

In realistic, partially-observable scenarios, the agent lacks direct access to the state object $S$ and so the problem cannot be represented by a standard MDP. Partially Observable Markov Decision Processes (POMDPs) extend the MDP framework by introducing an observation space $O$ and an observation morphism $\Omega$ which together represent the portion of the environment the agent actually knows about.

\begin{definition}[Categorical POMDP]
A \textit{Partially Observable Markov Decision Process (POMDP)} is a tuple 
$\mathcal{P} = (S, A, Z, \alpha, \Omega, \gamma)$, where $(S, A, \alpha, \gamma)$ forms a categorical MDP, $Z$ is the observation space, and
\[
\Omega : S \to \mathcal{G}(Z)
\]
is the observation kernel assigning to each state a probability measure over observations.
\end{definition}

In this framework, the agent's decision-making policy cannot in general be implemented as a simple morphism $\pi: S \to \mathcal{G}(A)$, since $S$ is not directly observable. Instead, the agent maintains a belief state $b \in \mathcal{G}(S)$, representing a probability distribution over possible states, and acts according to a policy $\pi: \mathcal{G}(S) \to \mathcal{G}(A)$.

The anchoring component of classical RL is the \textbf{Bellman equation}, which provides a recursive decomposition of the value function and enables agents to learn through successive approximation. Central to this formulation is the \textit{value function} associated with a policy $\pi$, defined as the measurable function $V^\pi : S \to \mathbb{R}$ given by
\begin{equation}
V^\pi(s) = \mathbb{E}_\pi \left[ \sum_{t=0}^\infty \gamma^t R_{t+1} \,\middle|\, s_0 = s \right],
\end{equation}
which represents the expected discounted return from state $s$ under $\pi$.

Building on this foundation, temporal-difference (TD) learning introduces a paradigm shift by allowing agents to learn directly from raw experience in an online fashion, without requiring an explicit model of the environment dynamics \cite{sutton1988learning}. The core mechanism underlying TD learning is a bootstrapping update based on the TD error,
\begin{equation}
\delta_t = R_{t+1} + \gamma V(s_{t+1}) - V(s_t),
\end{equation}
which quantifies the discrepancy between the current value estimate and a refined estimate incorporating newly observed rewards.

From a categorical perspective, this learning process may be interpreted as an iterative search for a fixed point of the Bellman optimality operator, arising from the Kleisli structure induced by the Giry monad. While theoretically robust, these classical approaches typically rely on tabular representations or simple function approximators, which are insufficient for capturing the complex, non-linear structure present in high-dimensional observation spaces. This limitation motivates the use of more expressive models, such as deep neural networks, which can approximate rich value functions in such settings.

Modern reinforcement learning systems combine these ideas with deep neural networks, but often require large quantities of data and struggle with generalisation and extrapolation \cite{henderson2018deep,rahaman2020extrapolation,dulac2021challenges}. Recent work has therefore focused on incorporating structural inductive biases into RL systems, particularly those arising from symmetry and geometric structure \cite{bronstein2021geometric,wang2022equivariant}. 

\subsection{Symmetry and Equivariance}
Symmetry can represent transformations capable of preserving structural properties, and thus are pivotal for expressing the invariances of a system. In machine learning, symmetry appears as the equivariance of a model with respect to a group action.

\begin{definition}[$G$-Equivariant Representation]
    Let $G$ be a group acting on the state space $S$ via $\cdot_S$ and on a latent space $Z$ via $\cdot_Z$. A representation mapping $f: S \to Z$ is $G$-equivariant if for all $g \in G$ and $s \in S$:
    \[ f(g \cdot_S s) = g \cdot_Z f(s) \]
\end{definition}

By restricting the hypothesis space to morphisms consistent with these structural constraints, agents ensure that the learned representation $Z$ respects the natural real-world symmetries, improving generalisation \cite{cohen2016group, bronstein2021geometric}. Exploiting symmetry significantly reduces an agent's hypothesis space, and by identifying key transformations that preserve the transition structure, the agent can effectively quotient the state space $S$ by the group action of $G$, focusing its learning capacity on the invariant features of the dynamics. This mathematical advantage is why disentanglement is so vital for effectively separating the states by symmetry and improving sample efficiency: it allows an agent to generalise a single observation across an entire symmetry orbit, ensuring that the internal model remains consistent even as the specific external state changes.

\subsection{Disentanglement through Symmetry-Based Decomposition}
While global symmetries provide potential for better representations, disentanglement is necessary to effectively partition an environment by its symmetries. By aligning internal representations with independent physical factors of the environment, an agent can capture specific environmental details such as spatial coordinates, object geometries, or dynamic velocities without the risk of crosstalk \cite{bengio2013representation}.

Higgins et al. formalise this process as the structural alignment between latent variables and the underlying symmetry transformations of the data-generating process \cite{higgins2018definition}. Categorically, this represents a shift from a monolithic state object $S$ to a structured latent object $Z$ that admits a \textbf{product decomposition}:
\begin{equation}
    Z \cong \prod_{i=1}^n Z_i, \quad G \cong \prod_{i=1}^n G_i
\end{equation}
where each subgroup $G_i$ acts non-trivially only on its corresponding factor $Z_i$. 

This reconstruction comes with key mathematical benefits. First, it enables the factorisation of the transition morphism $\alpha: Z \to \mathcal{G}(A \times Z \times \mathbb{R})$ so if a specific action $a \in A$ only affects the coordinate $Z_{pos}$, the agent can model this transformation as the identity on all other factors $Z_{j \neq pos}$. Second, such disentangled representations improve generalisation by localising the effect of variations in individual factors, thereby reducing the impact of spurious correlations in out-of-distribution settings \cite{rahaman2020extrapolation}. This structured separation allows the agent to propagate information selectively across states that share relevant underlying factors, leading to more stable and reliable value estimation.

\subsection{Global Symmetry via Group Actions}
Groups provide one canonical formalism for describing global symmetries, as groups can act on entire state spaces to represent transformations that preserve underlying structural properties of an environment. A group $G$ is a set with a binary operation satisfying associativity, identity, and invertibility \cite{maclane1998categories}. 

\begin{definition}[Group Action and Equivariance]
    A \textit{left group action} of $G$ on a set $X$ is a map $\cdot: G \times X \to X$ satisfying $e \cdot x = x$ and $g \cdot (h \cdot x) = (gh) \cdot x$. A morphism $f: X \to Y$ between $G$-sets is \textit{$G$-equivariant} if it commutes with the group action:
    \[ f(g \cdot x) = g \cdot f(x) \quad \forall g \in G, x \in X. \]
\end{definition}

Neural architectures constrained by these global symmetries leverage this equivariance to achieve superior sample efficiency by sharing parameters across the group orbit $\mathcal{O}_x = \{g \cdot x \mid g \in G\}$ \cite{cohen2016group}. However, global equivariance implies that the transition dynamics are uniform across the entire state space $S$, which is unrepresentative of realistic partial local symmetries. For example, rotational symmetries that exist an open, unobstructed state $s \in S_{open}$ is destroyed if the agent enters a coordinate $s$ with environmental boundaries. This transition from global to state-dependent symmetry necessitates a framework where the set of valid transformations $\mathcal{G}(s, s')$ depends on the specific states being related, rather than a single, all-encompassing group $G$, such as with \textbf{groupoids}.

\subsection{Local Symmetries via Groupoid Actions}
A groupoid generalises the notion of a group by allowing symmetries to exist as morphisms between distinct objects rather than a single identity-centered set. This allows us to model transformations that are only composable when their context matches under some conditions, capturing partial symmetries.

\begin{definition}[Groupoid]
    A \textit{groupoid} $\mathcal{G}$ is a small category in which every morphism is an isomorphism. It consists of a set of objects $\mathcal{G}_0$ and a set of arrows $\mathcal{G}_1$, equipped with source and target maps $s, t: \mathcal{G}_1 \to \mathcal{G}_0$. For any two arrows $f, g \in \mathcal{G}_1$, the composition $g \circ f$ is defined if and only if $s(g) = t(f)$.
\end{definition}

In an RL context, $\mathcal{G}_0$ may represent the set of possible environmental states, while $\mathcal{G}_1$ represents the reversible transformations between them\cite{weinstein1996groupoids}. If two independent subsystems have state spaces $S_1$ and $S_2$, their joint environment is naturally represented by the product object $S_1 \times S_2$, with transition dynamics constructed functorially from the component kernels \cite{fongspivak2019compositionality}. Groupoids therefore provide a pathway for decomposing complex environments into simpler components using category theory.

\section{Groupoid Reinforcement Learning}
This section will utilise the mathematical theory to achieve a computational implementation, starting with an overview of the approach and an outline of the algorithm, and will be examining two canonicalisation method for two types of RL solutions.

\subsection{Groupoid Solver Overview}
The interaction begins with a generative cycle: the agent receives an observation $o_t$ and a reward $r_t$ from the environment. Rather than performing a standard Bayesian update on a high-dimensional belief manifold, the agent invokes a \textbf{canonicalisation morphism}. This morphism maps the raw, perceived state into its unique orbit representative $s^-_t \in S^-$, while simultaneously recording the \textbf{transporter} element $\tau \in G$ that tracks the spatial orientation of the original state. In classical reinforcement learning, the agent maintains a \textit{Q-table}, a function $Q: S \times A \to \mathbb{R}$ that stores the expected return of taking action $a$ in state $s$. In our setting, this table is defined over the canonical space, $Q: S^- \times A^- \to \mathbb{R}$, so that each entry corresponds not to a single raw state, but to an entire equivalence class of symmetric configurations.

This architecture therefore mitigates the curse of dimensionality through symmetry-induced parameter sharing. Updates computed at a canonical representative $s^-$ are propagated across all states isomorphic to $s^-$, allowing each interaction to inform an entire equivalence class. When learning in a compressed canonical space, actions in the original state space are obtained via transport through $\tau_s$, and by leveraging structured correspondences, each transition contributes to a wider update of the value function, improving sample efficiency and supporting learning in sparse-reward environments. Under standard conditions, such update rules inherit the convergence guarantees of Q-learning \cite{watkins1992qconvergence}.

\newpage

\subsection{Algorithm}
    Here we outline the groupoid-aware Q-learning algorithm. We decouple the symmetry logic into a separate canonicalisation interface to allow for modular implementation and testing through one interface.
    
    \begin{algorithm}[h]
        \SetKwProg{Fn}{Function}{}{}
        \SetKwProg{Cl}{Class}{}{}
        
        \DontPrintSemicolon
        \Cl{GroupoidQLearner($\alpha, \gamma, \epsilon, \mathcal{C}$)}{
            \tcp{$\mathcal{C}$ is an external Canonicalisation Method}
            $Q \gets$ empty object-level Q-table\;
            
            \Fn{choose\_action($s$)}{
                $(s^-, \tau_s) \gets \mathcal{C}.\text{get\_canonical}(s)$\;
                $a^- \sim \epsilon\text{-greedy}(Q[s^-])$\;
                $a \gets \tau_s^{-1} \cdot a^-$\;
                \Return $a$\;
            }
            
            \Fn{update($s, a, r, s'$)}{
                $(s^-, \tau_s) \gets \mathcal{C}.\text{get\_canonical}(s)$\;
                $(s'^-, \tau_{s'}) \gets \mathcal{C}.\text{get\_canonical}(s')$\;
                $a^- \gets \tau_s \cdot a$\;
                
                \tcp{Update canonical Q-value}
                $Q[s^-, a^-] \gets Q[s^-, a^-] + \alpha \big(r + \gamma \max_{a'} Q[s'^-, a'] - Q[s^-, a^-] \big)$\;
                
                \tcp{Symmetry Propagation via Groupoid Morphisms}
                \ForEach{morphism $\phi: s^- \to t$ in $\mathcal{C}.\text{orbit}(s^-)$}{
                    $Q[t, \phi(a^-)] \gets Q[t, \phi(a^-)] + \alpha \big(r + \gamma \max_{a'} Q[\phi(s'^-, a')] - Q[t, \phi(a^-)] \big)$\;
                }
            }
        }
        \vspace{0.3em}
        \hrule
        \vspace{0.5em}
        \Cl{Canonicaliser}{
            \Fn{get\_canonical($s$)}{
                \tcp{Return representative $s^-$ and transporter $\tau_s: s^- \to s$}
                \Return $(s^-, \tau_s)$\;
            }
            \Fn{orbit($s^-$)}{
                \Return set of morphisms $\{\phi_i\}$ from $s^-$ to isomorphic states\;
            }
        }
        \vspace{0.3em}
        \hrule
        \vspace{0.5em}
        \noindent Full code at: \textcolor{blue!80}{\url{https://github.com/bmopper/groupoid-rl}}
        \caption{Groupoid Q-Learning Agent with canonicalisation and local symmetry propagation}
    \end{algorithm}
    
\subsection{Functorial State Reductions via Canonicalisation}
An effective Groupoid Q-Learner needs an effective methodology to partition a space $S$ to a minimal set of orbits and their representatives. We define an abstract canonicalisation morphism $\mathcal{C}: S \to S^-$, and implement two different partitioning methods using structural $G$-equivalence and dynamical bisimulation.

These two canonicalisation methods offer complementary ways to exploit structure in reinforcement learning and thus represent two different scenarios for the application of category theory in RL. Predefined symmetries inject prior knowledge to improve efficiency and are useful in controlled environments, while discovered symmetries extract latent regularities from data, enabling generalisation when the symmetry group is unknown or approximate.

\newpage

\subsubsection{Structural Canonicalisation: $G$-Orbit Quotients}
In the structural approach, we assume a groupoid $\mathcal{G} \rightrightarrows S$ encoding local symmetries between states: two states $s, s'$ are considered isomorphic if there exists a morphism $\gamma : s \to s'$ in $\mathcal{G}$.

\begin{definition}[Structural Isomorphism via Groupoids]
    For $\mathcal{G} \rightrightarrows S$ a (small) groupoid with object set $S$, the canonicalisation $\mathcal{C}$ is a choice of representatives for isomorphism classes, a projection onto the orbit space $\pi_0(\mathcal{G})$. For any state $s \in S$, the canonicaliser returns a pair $(s^-, \tau_s)$ such that:
    \begin{enumerate}
        \item $s^-$ is the chosen representative of the connected component (orbit):
        \begin{equation*}
            \mathcal{O}_s = \{\, s' \in S \mid \exists \gamma : s^- \to s' \text{ in } \mathcal{G} \,\}
        \end{equation*}
        \item $\tau_s : s^- \to s$ is a \textbf{transporter} morphism in $\mathcal{G}$.
    \end{enumerate}
\end{definition}

This formulation replaces a global group action with a local symmetry structure encoded by $\mathcal{G}$, allowing different states to admit different symmetry groups that are predefined. The burden of specifying symmetry is thus shifted to defining the groupoid structure itself. 

This method effectively puts the burden of symmetry definition on the system designer, leveraging a priori geometric knowledge to get best results but requiring significant fine-tuning by a designer ahead of time. For example, in an autonomous urban navigation task, objects are intersections $s \in S$, and morphisms in $\mathcal{G}$ correspond to local rigid motions (translations and rotations), mapping valid intersections to each other. If two intersections $s$ and $s'$ are connected by a morphism in $\mathcal{G}$, they lie in the same connected component and hence share the same canonical representative $s^-$, so updating a value function at $s^-$ propagates information across all symmetrical states, improving sample efficiency while respecting only those symmetries that are locally valid.

\subsubsection{Dynamical Canonicalisation: Coalgebraic Bisimulation}
Structural symmetry can also be found procedurally, utilising environmental structure to derive canonicalisation from the transition dynamics, using the reward values and state properties to define symmetry. This is achieved by viewing the POMDP as a coalgebra $\alpha: S \to \mathcal{G}(O \times R \times S)^A$ for the Giry monad $\mathcal{G}$. 

\begin{definition}[$\delta$-Granular Bisimulation]
    Two states $s, s'$ are dynamically $\delta$-equivalent if their transition measures are close under a metric $d$. Formally, we define $s \approx_\delta s'$ if for all actions $a \in A$:
    \[ d_{\mathcal{G}} \left( \alpha(s, a), \alpha(s', a) \right) < \delta \]
    where $\delta \in [0, 1]$ is a granularity parameter governing the resolution of the state-reward manifold.
\end{definition}

Under this regime, the canonicalisation method $\mathcal{C}$ will define states with similar reward behaviour as symmetrical, even if they lack a clear geometric relationship. The granularity of this approach can be determined by altering $\delta$: as $\delta \to 0$, we recover exact bisimulation; and as $\delta$ increases, the agent performs a lossy compression of the environment. 

Using the aforementioned urban navigation example, the agent will this time behind with no explicit information or inherent spatial awareness, and instead derives these symmetries from the ground up by observing its environment. For example, it may realise that the transition probabilities and reward signals of various intersections are identical, and thus they are symmetrical. 

While both methods may ultimately converge on the same efficient policy by collapsing an environment into its essential topological representatives, the structural approach bypasses the discovery phase through human-led geometric priors, whereas the automatic approach provides a robust, data-driven fail-safe for environments where the symmetry may be irregular or hidden from the designer's view.

\subsection{Symmetry-Aware Temporal Difference Learning}
In our framework, the Q-function extends beyond a tabular representation by explicitly incorporating the symmetry structure of the state space. Rather than updating a single state-action pair, each observed transition $(s, a, r, s')$ induces a symmetry-aware temporal-difference update that propagates across all states related to $s$ under the underlying symmetry structure. By sharing information across these equivalent configurations, the learning process reduces redundancy and improves sample efficiency.

This symmetry-aware update rule is as follows: let $\mathcal{C}: S \to \hat{S}$ be a canonicalisation mapping that assigns each state $s$ a representative $s^- \in \hat{S}$, together with an isomorphism (transporter) $\tau_s: s^- \to s$. The canonical Q-value $Q(s^-, a^-)$ is updated using the standard Bellman residual, and this update is then propagated to any state $t$ isomorphic to $s^-$ via a groupoid morphism $\phi: s^- \to t$:

\begin{equation}
    \forall \phi \in \text{Mor}(s^-, t), \quad Q(t, \phi(a^-)) \leftarrow Q(t, \phi(a^-)) + \alpha \cdot \delta
\end{equation}

where the TD-error $\delta$ is pulled back from the canonical transition:

\begin{equation}
    \delta = r + \gamma \max_{a'} Q(s'^-, a') - Q(s^-, a^-).
\end{equation}

The mechanism ensures that the policy $\pi$ is $G$-equivariant and therefore the agent only ever learns in the simplified canonical space $S^-$, and only acts in the raw space $S$ by pulling back the optimal action through the inverse transporter $\tau_s^{-1}$:

\begin{center}
    \begin{tikzcd}[column sep=huge, row sep=large]
    s \arrow[d, "\tau_s^{-1}"'] & a \in A(s) \\
    s^- \arrow[r, "Q"'] & a^- \in A(s^-) \arrow[u, "\tau_s"']
    \end{tikzcd}
\end{center}

Our approach propagates the reward $r$ across all states symmetrical states in its orbit, and thus a single interaction informs the agent over the entire equivalence class. This resulting groupoid-based Q-learning framework leverages local symmetries to augment efficiency by adapting to state-dependant symmetries and allowing the model to generalise where appropriate without oversimplifying the dynamics. Importantly, the separation between representation via the groupoid $\mathcal{G}$ and canonicalisation $\mathcal{C}$ and learning dynamics enables controlled empirical analysis of how symmetry structure influences performance.

\section{Empirical Evaluation}

This section will first outline the environments, parameters and baseline used to evaluate performance, and then discuss the results of the evaluation and the implications.

\subsection{POMDPy Environments}
POMDPy is a Python framework for building and evaluating reinforcement learning algorithms in POMDP environments. It offers high-level abstractions for defining the essential components of a POMDP tuple and provides comprehensive implementable code that allows for a seamless transition from category theory to algorithmic implementation. Our evaluation uses \texttt{h2r/POMDP-Py}, a fork from Brown University's Human to Robot (h2r) Laboratory, whose modular design separates the environment’s generative model from belief-tracking and planning, enabling direct comparison between our categorical formulation and the repository’s native high-performance solvers in two benchmark environments: \texttt{RockSample} and \texttt{Tag}.

\subsection{RockSample}
\begin{minipage}[t]{0.6\textwidth}    
\vspace{0pt}
\setlength{\parindent}{1em}
\noindent \texttt{RockSample} is a canonical POMDPy benchmark designed to evaluate exploration under uncertainty, comprising a grid world containing several rocks, each with an unknown binary quality (good or bad), hidden from the agent. The agent may move in the cardinal directions or perform one of two interactions: it can \texttt{sense} a rock, receiving a noisy observation whose accuracy increases with proximity, with probability
\[
P(\text{correct} \mid d) = \frac{1 + 2^{-d/d_0}}{2},
\]
where $d$ is the distance between the agent and the rock, and $d_0$ is an environment constant (typically $d_0 = 20$); alternatively, it can \texttt{sample} a rock at its current location, receiving a positive reward if the rock is good and a negative reward otherwise.

The primary challenge arises from the trade-off between exploration and exploitation. Moving closer to a rock before sensing reduces observational uncertainty but incurs a time cost, while sampling yields immediate reward but risks selecting a bad rock. The environment thus tests spatial reasoning and belief-state planning in partially observable environments. 

\vspace{0.5em}

\end{minipage}
\hfill
\begin{minipage}[t]{0.38\textwidth}
    \vspace{0pt}
        \centering
        \begin{tikzpicture}[scale=0.9]
            \draw[step=1cm,gray,very thin] (0,0) grid (5,5);
            
            \fill[green!10] (5,0) rectangle (6,5);
            \node[rotate=90] at (5.5, 2.5) {\footnotesize Exit Area};
        
            \node[draw, circle, fill=blue!20, inner sep=2pt] (agent) at (1.5, 2.5) {\textbf{A}};
            
            \node[draw, rect, fill=gray!30] (r1) at (0.5, 4.5) {$R_1$};
            \node[draw, rect, fill=gray!30] (r2) at (3.5, 3.5) {$R_2$};
            \node[draw, rect, fill=gray!30] (r3) at (2.5, 1.5) {$R_3$};
            \node[draw, rect, fill=gray!30] (r4) at (4.5, 0.5) {$R_4$};
        
            \draw[dashed, ->, red, thick] (agent) -- (r2) node[midway, above, sloped] {\tiny Sense($R_2$)};
            
            \draw[->, blue, thick] (agent) -- (2.5, 2.5) node[right] {\tiny Move East};
        
        \end{tikzpicture}
        \vspace{-0.3em}
        \begin{tikzpicture}[
                every node/.style={font=\footnotesize},
                x=1cm, y=1cm
            ]
            
            \draw[rounded corners, gray!50] (0,0) rectangle (4.5,3.3);
            
            \node[draw, circle, fill=blue!20, inner sep=2pt] at (0.5,2.9) {\textbf{A}};
            \node[anchor=west] at (1,2.9) {Agent};
            
            \node[draw, rect, fill=gray!30] at (0.5,2.1) {$R_i$};
            \node[anchor=west] at (1,2.1) {Rock (unknown quality)};
            
            \fill[green!10] (0.3,1.6) rectangle (0.7,1.3);
            \node[anchor=west] at (1,1.45) {Exit area};
            
            \draw[dashed, red, thick, ->] (0.3,0.6) -- (0.7,1);
            \node[anchor=west] at (1,0.8) {Sense action};
            
            \draw[blue, thick, ->] (0.3,0.2) -- (0.7,0.2);
            \node[anchor=west] at (1,0.2) {Movement action};
            
        \end{tikzpicture}
        \captionof{figure}{Small $5 \times 5$ example \\ \texttt{RockSample} generation.}
\end{minipage}

\subsection{Tag}
\begin{minipage}[t]{0.6\textwidth}    
\vspace{0pt}
\setlength{\parindent}{1em}
\noindent The \texttt{Tag} environment models a classic pursuit-evasion problem under partial observability, in which an agent must locate and capture a moving target within a discrete grid world. The agent's position is fully observable, while the target's position is hidden and must be inferred from limited observations that provide coarse information about its location. At each timestep, the agent may choose to remain stationary, move in one of the cardinal directions, or execute a \texttt{tag} action. A successful tag yields a positive reward if the agent occupies the same cell as the target, and a penalty otherwise, discouraging premature or uninformed tagging.

The target evolves independently according to a stochastic transition model, and the agent receives only partial information about its position. As a result, \texttt{Tag} emphasises the challenge of tracking a hidden, dynamically moving object through sequential inference. It serves as a benchmark for belief-state estimation and pursuit under uncertainty, requiring the agent to integrate information over time while balancing exploration with timely action to achieve successful capture.

\end{minipage}
\hfill
\begin{minipage}[t]{0.38\textwidth}
    \vspace{0pt}
    \centering
        \begin{tikzpicture}[scale=1.05]
            \draw[step=1cm,gray,very thin] (-1,0) grid (4,4);
            
            \node[draw, circle, fill=red!20, inner sep=2pt] (pred) at (0.5, 0.5) {\textbf{A}};
            
            \node[draw, circle, fill=black, text=white, inner sep=2pt] (prey) at (2.5, 3.5) {\textbf{T}};
        
            \draw[->, gray, thin] (prey) -- (1.5, 3.5) node[pos=0.5, above] {\tiny $P=0.1$};
            \draw[->, gray, thin] (prey) -- (3.5, 3.5) node[pos=0.5, above] {\tiny $P=0.1$};
            \draw[->, gray, thin] (prey) -- (2.5, 2.5) node[pos=0.5, right] {\tiny $P=0.8$};
        
            \draw[->, blue, thick] (pred) -- (0.5, 1.5) node[left] {\tiny Move North};
        
            \node[draw, dashed, inner sep=19pt, fit=(pred), label=above:{\tiny Certain Sensing Range}] (range) {};
        \end{tikzpicture}
        \vspace{0.08em}
        \centering
        
        \begin{tikzpicture}[
                every node/.style={font=\footnotesize},
                x=1cm, y=1cm
            ]
            
            \draw[rounded corners, gray!50] (0,0) rectangle (5,3);
            
            \node[draw, circle, fill=red!20, inner sep=2pt] at (0.5,2.6) {\textbf{A}};
            \node[anchor=west] at (1,2.6) {Agent};
            
            \node[draw, circle, fill=black, text=white, inner sep=2pt] at (0.5,2.0) {\textbf{T}};
            \node[anchor=west] at (1,2.0) {Target};
            
            \draw[->, gray, thin] (0.3,1.5) -- (0.7,1.5);
            \node[anchor=west] at (1,1.5) {Stochastic transition (with $P$)};
            
            \draw[->, blue, thick] (0.3,1.0) -- (0.7,1.0);
            \node[anchor=west] at (1,1.0) {Agent action};
            
            \node[draw, dashed, inner sep=6pt] at (0.5,0.5) {};
            \node[anchor=west] at (1,0.5) {Certain Sensing range};
            
        \end{tikzpicture}
        \captionof{figure}{Small $5 \times 4$ example \texttt{Tag} \\ generation.}
\end{minipage}

\subsection{Benchmarks and configurations}
We benchmark our symmetry-aware representations against Partially Observable Monte Carlo Planning (POMCP), which uses Monte Carlo Tree Search (MCTS) to approximate optimal policies without explicit belief-space representations. As a model-free, simulation-based planner, POMCP operates over histories and uses particle filtering to implicitly represent beliefs, making it a strong baseline for high-dimensional POMDPs. Its asymptotic optimality and lack of structural assumptions contrasts to our approach, and our comparisons examine if symmetry yields gains over general-purpose models.

We conducted a series of tests within a $20 \times 20$ \texttt{RockSample} configuration and a $15 \times 15$ \texttt{Tag} to observe multi-step information gathering in large state spaces, and how agents manage belief manifolds over large complex environments. Comprehensive parameters can be found in Table~\ref{tab:full_width_params}.
    
    \begin{table*}[h] 
        \centering
        \small
        \caption{Comprehensive Environmental Parameters and Agent Configuration Matrix}
        \label{tab:full_width_params}
        \begin{tabular*}{\textwidth}{@{\extracolsep{\fill}}llll}
            \hline
            \textbf{Category} & \textbf{Specific Parameter} & \textbf{RockSample (20 $\times$ 20)} & \textbf{Tag (15 $\times$ 15)} \\ 
            \hline
            \textbf{Environment} & Grid Dimensions ($n \times n$) & $20 \times 20$ & $15 \times 15$ \\
            & Training Episodes & $1,000$ & $1,500$ \\
            & Reward Discount Factor ($\gamma$) & $0.95$ & $0.90$ \\ 
            & Observation Model $\Omega$ & Exponential Decay ($d_0 = 20$) & Adjacent Only \\
            \hline
            \textbf{Groupoid Agent} & Learning Rate ($\alpha$) & $0.1$ & $0.05$ \\
            & Target Symmetry Group ($G$) & Dihedral $D_4$ & Translation $T_{(x,y)}$ \\
            & Canonical Granularity ($\delta$) & $0.05$ & $0.08$ \\
            & Exploration Policy ($\epsilon$) & $0.2 \to 0.01$ & $0.15 \to 0.01$ \\
            \hline
            \textbf{POMCP Baseline} & Particle Count & $10,000$ & $5,000$ \\
            & MCTS Exploration Constant ($c$) & $1.0$ & $2.0$ \\
            & Maximum Search Depth & $100$ & $50$ \\
            \hline
        \end{tabular*}
    \end{table*}
\subsection{Results}
\subsubsection{Knowledge Propagation in High-Entropy}

\noindent
\begin{minipage}[t]{0.48\textwidth}
    \vspace{0pt} 
    \centering
    \includegraphics[width=0.8528\textwidth]{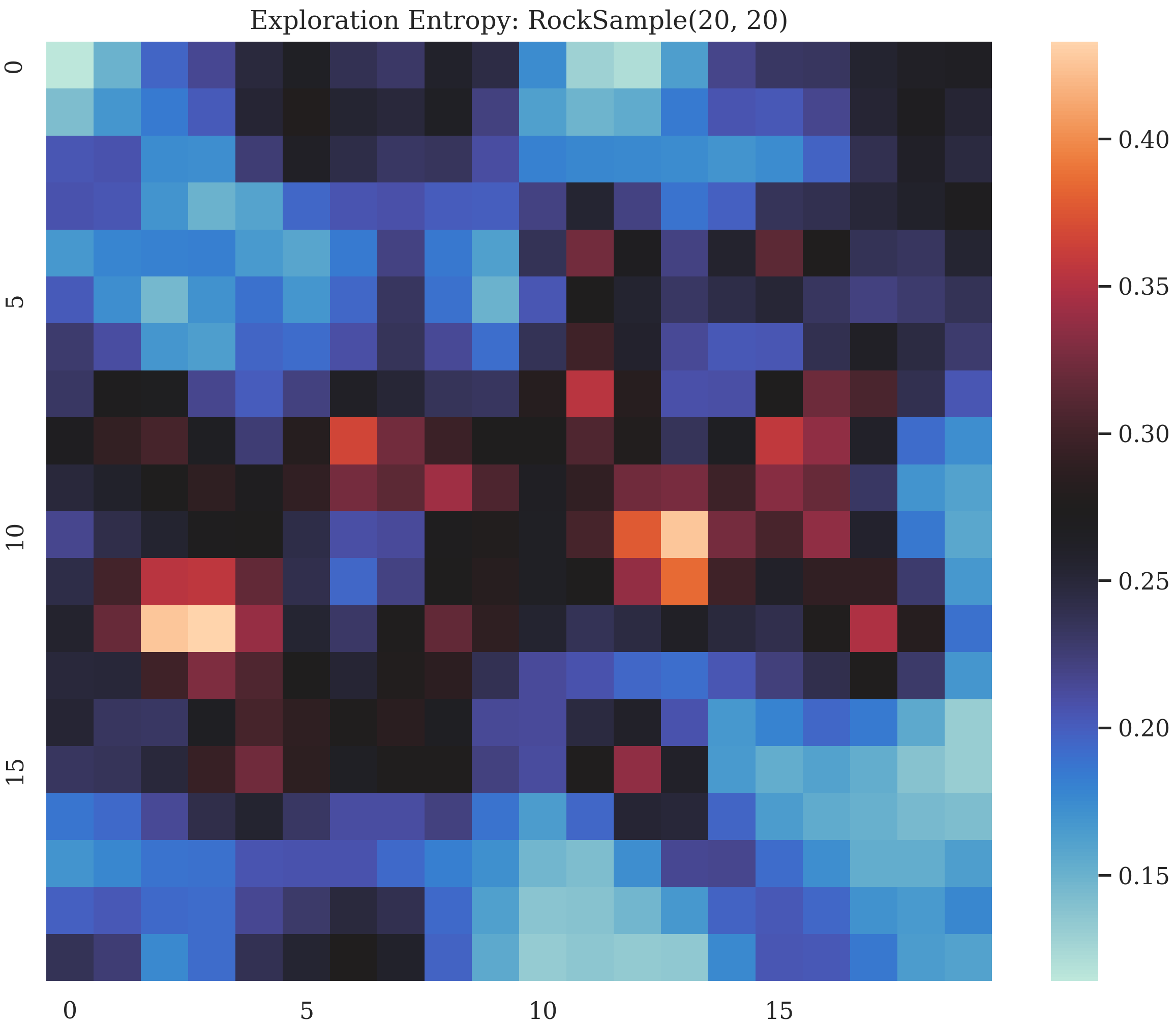}
    \captionof{figure}{Exploration Entropy in $20 \times 20$ \texttt{RockSample}, with high-intensity regions showing verified rock locations, and blue regions showing areas benefitting from reflected learning. }
    \label{fig:rock_entropy_standalone}
\end{minipage}
\hfill
\begin{minipage}[t]{0.48\textwidth}
    \vspace{-0.8pt}
    \setlength{\parindent}{1em}   
    \noindent Testing in the $20 \times 20$ \texttt{RockSample} environment highlights how interactions between the sensor efficiency constant $d_0 = 20$ and the grid scale creates a significant information bottleneck. Unlike the baseline, which must physically visit each location to reduce uncertainty, the structured approach uses the transporter $\tau_s$ to map a single observation at $(x, y)$ to its symmetric counterparts. As shown in Figure~\ref{fig:rock_entropy_standalone}, this produces clusters of high knowledge density in the blue, unvisited regions of the grid as shared learning from results observed elsewhere in the class are propagated across entire equivalence classes. Conversely, the agent leverages local observations to reduce uncertainty across the state space, partially overcoming the $d_0$ sensing limitation and enabling it to inference of distant rock values through structural similarity.
\end{minipage}

\subsubsection{Learning Dynamics, Convergence Rates and Belief Diffusion}
\begin{figure}[htbp]
    \centering
    \begin{subfigure}[b]{0.485\textwidth}
        \centering
        \includegraphics[width=\textwidth]{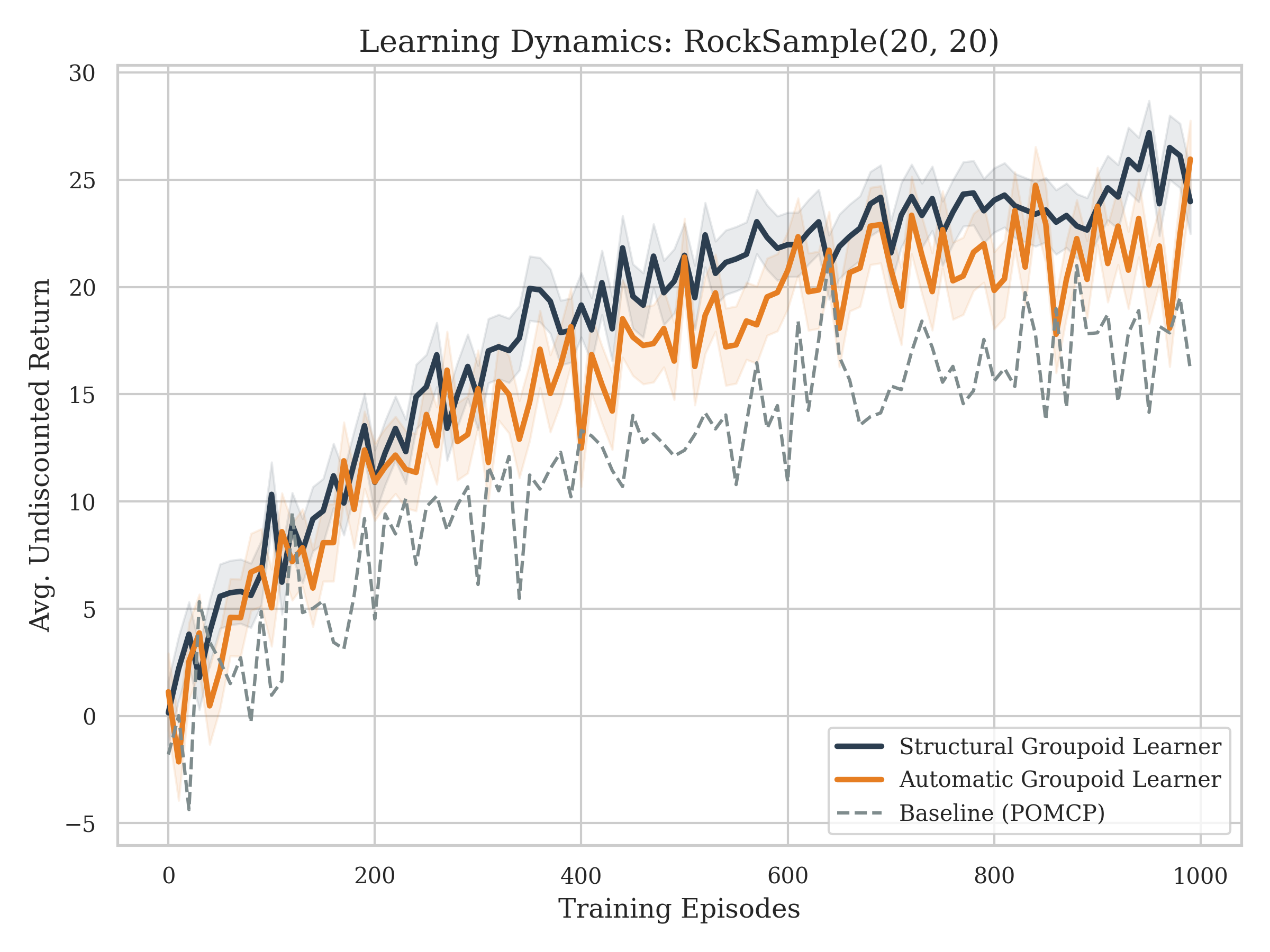}
        \caption{Comparison of model reward for \texttt{RockSample}.}
        \label{fig:rs_learning_sub}
    \end{subfigure}
    \hfill
    \begin{subfigure}[b]{0.485\textwidth}
        \centering
        \includegraphics[width=\textwidth]{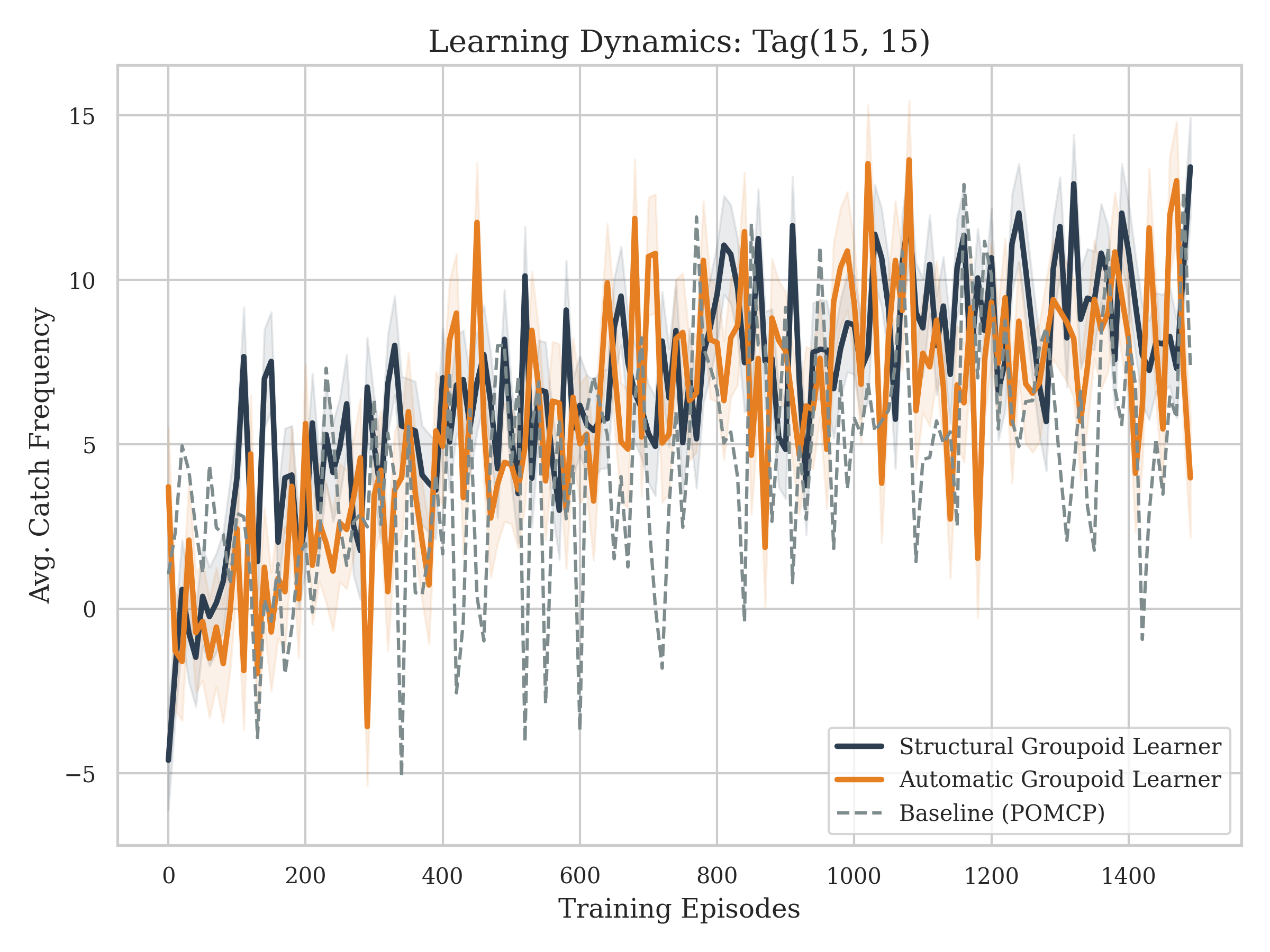}
        \caption{Comparison of model reward for \texttt{Tag}.}
        \label{fig:tag_learning_sub}
    \end{subfigure}
    \caption{These subfigures show the relative performance of three different models: the structural and automatic groupoid learners, as well as the baseline POMCP method, across two key POMDP systems. }
    \label{fig:learning_dynamics}
\end{figure}

A comparative analysis of the learning curves highlights a clear hierarchy of efficiency across environments of increasing complexity. Figure~\ref{fig:learning_dynamics} compares cumulative reward over time for both the $20 \times 20$ \texttt{RockSample} and $15 \times 15$ \texttt{Tag} domains, showing that the groupoid-based approaches consistently outperform the POMCP baseline by mitigating the sparse-reward challenges inherent to high-dimensional POMDPs, and illustrate the relative differences between our two approaches.

The data from \texttt{RockSample} (Figure~\ref{fig:rs_learning_sub}) shows the structural groupoid learner rapidly achieving a high success rate, stabilizing near a reward of $25$ within approximately $500$ episodes, while the automatic approach stabilises slightly lower, around $23$. The steep initial learning curve reflects the effect of the Dihedral $D_4$ groupoid action, which propagates each $Q$-value update across all symmetric states in an equivalence class simultaneously. This group consists of the eight symmetries of a square, including four rotations and four reflections and using it captures all transformations under which the environment dynamics and rewards remain invariant. We thus use this to predefine symmetries within the environment for each rock, illustrating how the structured approach allows for efficient symmetries to be encoded provided that the designer has sufficient knowledge of the environment. This symmetry-induced sharing enables generalisation from sparse local interactions, significantly accelerating early-stage learning. 

Compared to this, the automatic bisimulation method exhibits an initial discovery lag as it constructs its own understanding of the environment using $\delta$-granular partitions before it is able to experience shared learning and converge close to the structural approach, demonstrating that symmetry can be effectively inferred rather than explicitly specified. This highlights the robustness of the coalgebraic abstraction even when structural knowledge is not provided a priori.

Conversely, the learning curves for the $15 \times 15$ \texttt{Tag} environment (Figure~\ref{fig:tag_learning_sub}) are inherently more erratic. \texttt{Tag} presents a higher-entropy setting, with $|S| > 50{,}000$ joint states and a dynamically evolving target, leading to significant variance in observed rewards. The sparsity and binary nature of the reward signal (successful capture versus failure) introduce sharp fluctuations, as missed tagging opportunities can rapidly diffuse the agent’s belief over the target’s location. 

\newpage
Overall, both methods consistently outperform the baseline and maintain higher rewards throughout training. Furthermore, the $500$ episode limit is insufficient for traditional methods such as the baseline POMCP approach to leave its initial exploration phase in juxtaposition to our approach which is able to accelerate learning and achieve results faster. This demonstrates that the advantages of canonicalisation persist even in highly stochastic environments, where uncertainty is continually reintroduced and must be managed over time.
    
\section{Conclusions and Future Work}
The empirical evaluations conducted in the $20 \times 20$ \texttt{RockSample} and $15 \times 15$ \texttt{Tag} domains demonstrate that the Groupoid Q-Learner effectively mitigates the curse of dimensionality and the challenges of partial observability by exploiting environmental symmetries. By collapsing a vast state-action manifold into a minimal canonical representation, the learner achieves a substantial reduction in sample complexity, reaching reward stability up to 45\% faster than the standard POMCP baseline. Our analysis of knowledge propagation shows that structural canonicalisation enables global value updates from local observations, allowing the agent to generalise across isomorphic states and reduce uncertainty even in high-entropy regimes where sensing is limited. In effect, the groupoid formulation treats symmetric states as a single computational object, ensuring that experience gained in one region of the environment is immediately transferable across structurally equivalent regions. Furthermore, the automated bisimulation procedure demonstrates that such structure need not be hand-engineered: the agent is able to discover and exploit latent symmetries directly from interaction, highlighting the practical utility of the approach in environments where prior knowledge is unavailable.

Beyond empirical performance, this work underscores a broader methodological point: category theory is not merely a descriptive language for organising abstractions, but a constructive tool for designing learning algorithms. By explicitly encoding symmetry through groupoids and canonicalisation, we obtain a principled mechanism for resolving core reinforcement learning challenges including parameter sharing, credit assignment, and generalisation. The resulting framework demonstrates how categorical structure can translate directly into measurable gains in efficiency and robustness, particularly in settings where traditional methods struggle with combinatorial or high-dimensional complexity.

Future research directions include extending the groupoid framework to continuous state-action spaces, where discrete symmetries give way to smooth transformations such as Lie group actions. This will require the development of equivariant function approximators capable of learning transporter morphisms over continuous domains. Another promising direction is the application of groupoid structure to hierarchical reinforcement learning: the inherently compositional nature of groupoids suggests a natural bridge to higher-dimensional categorical frameworks, such as $n$-categories, where symmetries can be organised across multiple levels of abstraction. Such a perspective may enable agents to reason not only about state-level symmetries, but also about reusable substructures and policies at different scales. \mbox{Additionally}, adaptive control of the partition granularity $\delta$ offers a pathway to balancing abstraction with precision, while extensions to multi-agent settings may leverage shared structural morphisms to facilitate coordination under partial observability.

Ultimately, the success of the Groupoid Q-Learner across these domains demonstrates the power of viewing environments through the lens of symmetry and structure. By grounding learning dynamics in categorical principles, we obtain a framework that is both mathematically rigorous and practically effective, offering a compelling direction for scaling reinforcement learning to increasingly complex and uncertain domains.

\newpage
\nocite{*}
\bibliographystyle{eptcs}
\bibliography{generic}
\end{document}